\documentclass[conference]{IEEEtran}
\def\IEEEtitletopspace{18pt}

\usepackage{cite}
\usepackage{amsmath,amssymb}
\usepackage{graphicx}
\usepackage{xcolor}
\usepackage{microtype}
\usepackage[inkscapelatex=false]{svg}

\usepackage{tikz}
\usetikzlibrary{shapes.geometric, shapes.misc, positioning, calc, fit, backgrounds, arrows.meta}

\usepackage{tabularray}

\usepackage[hidelinks]{hyperref}
\usepackage[noabbrev]{cleveref}

\newcommand{\bs}[1]{\boldsymbol{#1}}

\usepackage{tikz}
\newcommand\copyrighttext{%
  \footnotesize \textcopyright 2025 IEEE.  Personal use of this material is permitted.  Permission from IEEE must be obtained for all other uses, in any current or future media, including reprinting/republishing this material for advertising or promotional purposes, creating new collective works, for resale or redistribution to servers or lists, or reuse of any copyrighted component of this work in other works.}
\newcommand\copyrightnotice{%
\begin{tikzpicture}[remember picture,overlay]
\node[anchor=south,yshift=10pt] at (current page.south) {\fbox{\parbox{\dimexpr\textwidth-\fboxsep-\fboxrule\relax}{\copyrighttext}}};
\end{tikzpicture}%
}
\definecolor{blue}{rgb}{0, 0.4470, 0.7410}
\definecolor{orange}{rgb}{0.8500, 0.3250, 0.0980}
\definecolor{yellow}{rgb}{0.9290, 0.6940, 0.1250}
\definecolor{purple}{rgb}{0.4940, 0.1840, 0.5560}
\definecolor{green}{rgb}{0.4660, 0.6740, 0.1880}
\definecolor{lightblue}{rgb}{0.3010, 0.7450, 0.9330}
\definecolor{red}{rgb}{0.6350, 0.0780, 0.1840}

\newcommand{\tikzConcept}{
  \begin{tikzpicture}[%
        rounded corners=0.3mm,
        >={Stealth[round]},
        every node/.style={font=\large},
        trapezium/.style={
            shape=trapezium, 
            trapezium stretches=true,
            draw=blue, fill=blue!20, 
        },
    ]
    \node (x) {$\bs x$};
    
    \node[%
        right=of x, 
        trapezium, 
        shape border rotate=270,
        minimum width=15mm, 
        minimum height=10mm, 
        trapezium angle=70, 
    ] (phi) {$\Phi$};
    \node[%
        right=2mm of phi, 
        draw=green, fill=green!20, 
        minimum height=9mm, 
    ] (z) {$\bs z$};
    \node[%
        right=2mm of z, 
        trapezium, 
        shape border rotate=270, 
        minimum width=7mm, 
        minimum height=10mm, 
        trapezium angle=70,
    ] (psi) {$\Psi$};
    
    \node[
        right=of psi,
    ] (y) {$\bs y$};
    
    \node[%
        below=of z,
        draw=yellow, fill=yellow!30, 
        text width=20mm, 
        align=center,
        inner sep=2mm,
        font=\footnotesize
    ] (monitor) {Runtime Safety\\ Monitor};
    
    \node[below right=3mm and 1mm of monitor, align=center, font=\footnotesize] (alert) {Safety Alert};
    
    \draw[->] (x) -- (phi);
    \draw[] (phi) -- (z);
    \draw[] (z) -- (psi);
    \draw[->] (psi) -- (y);
    \draw[->] (x) |- (monitor);
    \draw[->] (y) |- (monitor);
    \draw[->] (z) -- (monitor);
    \draw[->] (monitor) |- (alert);
    
    \begin{pgfonlayer}{background}
      \node[
        draw=lightgray,
        fill=lightgray!10,
        dashed,
        fit=(phi)(z)(psi),
        inner sep=4mm,
        label={[text=gray, font=\footnotesize,]above:{Monitored Neural Network}}
      ] (fitbox) {};
    \end{pgfonlayer}
  \end{tikzpicture}
}

\begin{document}
\title{Runtime Safety Monitoring of Deep Neural Networks for Perception: A Survey}

\author{
    \IEEEauthorblockN{%
        Albert Schotschneider\IEEEauthorrefmark{1},
        Svetlana Pavlitska\IEEEauthorrefmark{1}\IEEEauthorrefmark{2} and
        J. Marius Zöllner\IEEEauthorrefmark{1}\IEEEauthorrefmark{2}
    }
    \IEEEauthorblockA{
        \IEEEauthorrefmark{1}
        Department of Technical Cognitive Systems\\
        FZI Research Center for Information Technology, Karlsruhe, Germany\\
        \texttt{\{schotschneider, pavlitska\}@fzi.de}
    }
    \IEEEauthorblockA{
        \IEEEauthorrefmark{2}
        Institute for Applied Informatics and Formal Description Methods\\
        Karlsruhe Institute of Technology (KIT), Karlsruhe, Germany\\
        \texttt{marius.zoellner@kit.edu}
    }
}


\maketitle
\copyrightnotice

\begin{abstract}
    Deep neural networks (DNNs) are widely used in perception systems for safety-critical applications, such as autonomous driving and robotics. However, DNNs remain vulnerable to various safety concerns, including generalization errors, out-of-distribution (OOD) inputs, and adversarial attacks, which can lead to hazardous failures. This survey provides a comprehensive overview of runtime safety monitoring approaches, which operate in parallel to DNNs during inference to detect these safety concerns without modifying the DNN itself. We categorize existing methods into three main groups: Monitoring inputs, internal representations, and outputs. We analyze the state-of-the-art for each category, identify strengths and limitations, and map methods to the safety concerns they address. In addition, we highlight open challenges and future research directions.
\end{abstract}


\section{Introduction}\label{sec:introduction}

Deep neural networks (DNNs) have become the backbone of perception systems in safety-critical
applications such as autonomous driving, robotics, and industrial automation. Their ability to
process high-dimensional data has enabled advancements in object detection, semantic segmentation,
and scene understanding. However, despite their success, DNNs remain vulnerable to functional
insufficiencies: unpredictable failures caused by generalization errors, out-of-distribution (OOD)
inputs, and adversarial attacks. These vulnerabilities pose significant risks in real-world
deployments, where a single misclassification or missed detection could lead to catastrophic
consequences.
Traditional approaches to ensuring DNN reliability, such as testing or adversarial training, often
fall short in dynamic environments. Testing cannot exhaustively cover all edge cases while
retraining models to address new failure modes is impractical for deployed systems. Runtime safety
monitors (RSMs) offer a complementary solution by continuously observing the DNN's operation in
parallel during inference and detecting safety concerns without modifying the monitored DNN.

RSMs are a powerful mechanism to address the generalization and adaptability issues of DNNs. While RSMs cannot replace rigorous offline testing of safety-critical systems, they can complement traditional approaches ensuring safety assurance and thus making DNNs more robust and trustworthy.

\begin{figure}[t]
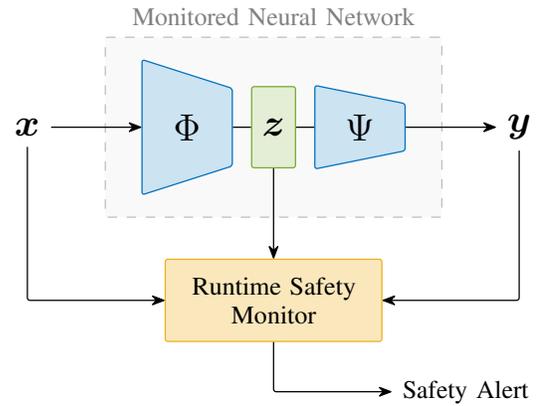

    \centering
    \scalebox{1.2}{\tikzConcept}
    \caption{%
        A system architecture for monitoring a DNN using a runtime safety monitor, which can observe inputs $\bs x$, internal layers $\bs z$, or outputs $\bs y$ of the monitored DNN to detect safety concerns and trigger a safety alert. The neural network component $\Phi$ transforms the input $\bs x$ into a latent representation $\bs z$, while component $\Psi$ of the system may handle diverse tasks, e.g., classification, segmentation, or decoding.
    }
    \label{fig:concept}
\end{figure}

This survey provides a comprehensive overview of runtime safety monitoring approaches for DNN-based
perception systems. We focus on methods that:

\begin{enumerate}
    \item operate externally to the monitored DNN,
    \item do not modify the monitored DNN, and
    \item can observe inputs, internal representations, or outputs of the monitored DNN.
\end{enumerate}

Such non-intrusive monitoring methods preserve certification and compliance of DNNs, avoid retraining, and can provide plug-and-play compatibility.
A system architecture of such a runtime safety monitor observing a DNN is shown in~\Cref{fig:concept}, where a runtime safety monitor can observe inputs, internal representations, or outputs of the monitored DNN, and trigger a safety alert if a safety concern is detected. In this work, we propose a categorization of monitoring approaches and map them to DNN-related safety concerns.

\newpage
\section{Previous Surveys}\label{sec:previous}

Methods to ensure the safety of DNNs were a focus of several previous works~\cite{houben2022inspect,muhammad2020deep,mohseni2023taxonomy}, while most approaches are applied during DNN training and design. Huang et al.~\cite{huang2020asruvey} provide a comprehensive survey of methods ensuring the safety and trustworthiness of DNNs, focusing on the design-time approaches, including formal verification and coverage testing. In this work, we concentrate on inference-time methods to ensure DNN safety.

The survey by Rahman et al.~\cite{rahman2021runtime} focuses on the perception of mobile robots and groups existing RSM approaches into three categories: monitoring based on (1) past experiences, (2) inconsistencies during inference, and (3) uncertainty estimation and confidence. RSMs of the first group are auxiliary networks trained on past positive and negative examples to predict the output of the monitored DNN. The methods from the second group monitor inconsistencies during inference, and those from the third group use uncertainty estimation and confidence. Only the second group's methods thus overlap with the focus of our survey. Additionally, Rahman et al. categorize papers based on the position in the robotic perception pipeline where the monitoring is performed. Their overview demonstrates that more methods rely on output evaluation. They further consider group methods based on supervision with examples of success and failure during training. In this work, we stick to a narrower view of RSMs, which do not require training-time data and do not modify the monitored DNN.

Differently from surveys that categorize existing works into groups, Ferreira et al.~\cite{ferreira2023runtime,ferreira2024safety} instead start with the identification of safety considerations and focus on the way RSMs are integrated into the system. Regarding the latter, the approaches are categorized into external, which operate independently from the monitored model, and internal, which are part of the monitored model. Internal approaches include uncertainty estimation, incorporating domain knowledge, and learning with rejection. External methods include monitoring inputs (input reconstruction and signal processing), monitoring internal layers (coherence, binary activations, continuous values), monitoring outputs (input perturbations, ensemble methods, consistency checks, softmax confidence),  and using external sensors. In this work, we focus only on the external RSMs, which allow for easier compatibility and deployment.

A safeguard (or guard) is a closely related term, referring to a mechanism preventing a DNN from making unsafe decisions or operating outside acceptable boundaries. Unlike an RSM, it typically doesn't require full introspection of the DNN’s internals but acts as a protective barrier around or in front of the DNN’s output or input~\cite{hussain2022deepguard,wang2020dnnguard}.

A further related field is DNN introspection, i.e., examining and analyzing the internal states or structure of a DNN, such as neuron activations, layer outputs, attention maps, or learned weights. Unlike an RSM, DNN instrospection is primarily used for understanding, debugging, or interpreting the model~\cite{yatbaz2024introspection}. The information gained via DNN introspection can, however, be used in safety ensurance mechanisms.

\newpage
\section{DNN-related Safety Concerns}\label{sec:safety_concerns}

To cluster methods, we adopt the ISO 21448 (SOTIF) definition of functional insufficiency, which refers to "an insufficiency inherent in the system that may lead to hazards"~\cite{sotif}. For DNN-related safety concerns, we follow Willers et al.~\cite{willers2020safety}, who define these as "underlying issues that may negatively affect the safety of a system." Their framework identifies nine key concerns, including data distributions that inadequately approximate real-world conditions, brittleness of DNNs under edge-case scenarios, and insufficient integration of safety considerations into performance metrics.

\subsection{Generalization Errors}

Generalization errors or in-distribution errors refer to incorrect predictions made by a deep neural network on inputs that are drawn from the same distribution as the training data~\cite{Jin2019QuantifyingTG, Jakubovitz2018GeneralizationEI}. These errors arise due to the model's insufficient capacity to represent the underlying data distribution or overfitting to training data noise and biases.
Overfitting is one of the primary causes of generalization errors.
When a DNN learns to focus on noisy or irrelevant patterns in the training data, it compromises its ability to adapt to new inputs. Another factor contributing to generalization errors is the lack of diversity in the training data. If a dataset does not adequately represent real-world input variations, the DNN fails to develop robust feature representations.

\subsection{Out-of-Distribution Errors}

Out-of-distribution errors occur when a DNN makes incorrect predictions on input data that are drawn
from a different distribution than the training data~\cite{hendrycks2016baseline}. Examples of OOD
inputs include new object classes, unusual poses or viewpoints, different image styles or
textures~\cite{liang2017enhancing}. Since the neural network has not learned the features to
robustly handle such inputs, it tends to make over-confident yet incorrect predictions. While DNNs
excel at learning patterns from the training data, their inability to recognize inputs from unseen
distributions exposes a fundamental limitation of their generalization capabilities.

Out-of-distribution inputs can take various forms. For example, in image classification tasks, OOD
data may include new object classes that were not present during training. In such cases, the DNN is
likely to produce high-confidence predictions that are completely erroneous, posing significant
safety risks.
The primary cause of OOD errors lies in the DNN's inability to distinguish between data it has
learned and data it has never seen. Since the network's feature extraction is inherently tied to the
training distribution, it struggles to generalize when inputs fall outside this distribution. Rather
than recognizing uncertainty, the model extrapolates its learned patterns, leading
to unreliable predictions.

\subsection{Adversarial Attacks}

Deep neural networks are inherently vulnerable to deliberately generated noise patterns (adversarial inputs)~\cite{Szegedy2013IntriguingPO, Goodfellow2014ExplainingAH}. These attacks can be performed both in digital and real-world settings. For the latter, visible adversarial noise can be concentrated in a specific image area in an adversarial patch. Adversarial patch attacks were also successfully shown for different perception tasks, in particular object detection~\cite{pavlitskaya2022suppress, Pavlitskaya2022AdversarialVO}, semantic segmentation, traffic sign
detection and recognition~\cite{pavlitska2023adversarial}, and steering angle predictions~\cite{Pavlitskaya2020FeasibilityAS}.

Mitigation strategies for adversarial attacks include adversarial training~\cite{Madry2017TowardsDL}, input processing, robust model architectures, and defensive distillation. Furthermore, adversarial attacks can be detected using statistical anomaly detection, OOD detection, and gradient-based analysis, and activation patterns can be monitored for inconsistencies.

\section{Monitoring Approaches}\label{sec:approaches}
In the following section, we describe three major groups of safety monitoring approaches, which are monitoring inputs, internal representations, and outputs.
To assess which DNN-related safety concerns are addressed by various safety monitoring approaches, we first perform a quantitative analysis of existing publications (see \Cref{tab:overview}).

\begin{table*}[tbp]
    \centering
    \caption{%
      Overview of state-of-the-art monitoring approaches of deep neural
      networks. This table details each method's authors, publication
      year, reference, specific task addressed,
      architecture of the monitored neural network, datasets utilized, components being monitored, and the
      events captured.
    }
    \begin{tblr}{%
        colspec={Q[c]Q[c]Q[c]X[c]X[c]X[c]X[c]X[c]},
        rows = {m},
        row{1} = {font=\bfseries},
        hline{1,Z} = {1pt},
        hline{2} = {0.5pt},
        row{even} = {bg=lightgray!20},
    }
    Author & Year & Ref. & Task & Architecture & Dataset & What is monitored? & Events captured \\

    Cheng et al. & 2019 & \cite{cheng2019runtime} & Image Classification & CNN & MNIST, GTSRB & Neuron activation patterns & OOD \\
    Henzinger et al. & 2019 & \cite{Henzinger2019OutsideTB} & Image Classification & CNN & MNIST, CIFAR-10, GTSRB & Hidden layers & OOD \\
    Lukina et al. & 2020 & \cite{Lukina2020IntoTU} & Image classification & VGG16, Custom CNNs & MNIST, CIFAR-10, GTSRB & Neuron activation values at feature layers & Novel classes at inference time \\
    Shao et al. & 2020 & \cite{Shao2020IncreasingTT} & Image classification & VGG16 & CIFAR-10, CINIC-10, STL-10, ImagetNet2012, GTSD & Accuracy of target DNN & Correct/Incorrect classification \\
    Hell et al. & 2021 & \cite{Hell2021MonitoringPR} & Depth estimation & U-Net, VAE, SSD & CARLA, DrivingStereo & OOD scores from VAE, LR, and SSD & Distributional shifts due to fog and rain \\
    Kantaros et al. & 2021 & \cite{kantaros2021real} & Detection of adversarial attacks on images & ResNet-56, ResNet-50, LISA-CNN, GTSRB-CNN & MNIST, CIFAR-10, ImageNet, AdvNet & Softmax outputs & Digital and physical adversarial inputs \\
    Hashemi et al. & 2022 & \cite{Hashemi2022RuntimeMF} & Object detection & PolyYOLO & Cityscapes & Neuron activation values & OOD \\
    Roy et al. & 2022 & \cite{roy_runtime_2022} & Image classification Object detection & MLP, Graph Markov Neural Network & CIFAR-10, SVHN, COCO-OOC & Joint distribution of inputs, outputs, and attribution. Spatial and temporal relationships & OOD inputs and novel classes and out of context objects \\
    Klingner et al. & 2022 & \cite{klingner2022detecting} & Depth estimation Semantic segmentation & ResNet18 & Cityscapes, KITTI & Edge consistency between input image, depth output, and segmentation output & Adversarial perturbations \\
    Giessler et al. & 2023 & \cite{Geissler2023ALS} & Image classification Object detection & YOLOv3, RetinaNet, ResNet50, AlexNet & COCO, KITTI & Activations of CNN layers & Silent data corruption \\
    Kaur et al. & 2023 & \cite{kaur_predicting_2023} & Image classification Object detection & ResNet18, ResNet34, DenseNet & GTSRB, ONCE & Conformance between DNN classifiers & OOD \\
    Hacker et al. & 2023 & \cite{hacker2023insufficiency} & Traffic sign recognition & 5-layer CNN, YOLOv2 & GTSDB, GTSRB & Adversarial perturbations, confidence scores, and saliency maps & OOD inputs, adversarial perturbations, in-distribution errors \\
    Liu et al. & 2023 & \cite{Liu2023InputVF} & Image classification & MLP, CNN & MNIST, CIFAR-10 & Local robustness radius of input features & Adversarial samples \\
    Asad et al. & 2024 & \cite{asad2024beyond} & Image classification & Adversarial Autoencoder (AAE) & MNIST, Fashion-MNIST, Coil-100 & Reconstruction error and latent space distribution & OOD inputs \\
    Yatbaz et al. & 2024 & \cite{Yatbaz2024RuntimeMO} & 3D object detection & PointPillars, CenterPoint & KITTI, NuScenes & Activation patterns from early, mid, and final backbone layers & Missed object detections \\
    Wu et al. & 2024 & \cite{Wu2024BAMBA} & Object detection & ResNet-50, ResNet-101 & KITTI, BDD100K, COCO & Feature activations at fully connected layers & OOD inputs \\
    \end{tblr}
    \label{tab:overview}
\end{table*}

\begin{table*}[htbp]
  \centering
    \caption{%
        Mapping of DNN-related safety concerns to the major safety monitoring approaches. The table highlights which methods address specific concerns, providing an overview of the current state-of-the-art solutions.
    }
    \begin{tblr}{%
        hline{1} = {2-4}{0.5pt},
        hline{2,Z}   = {0.5pt},
        vline{1} = {2-Z}{0.5pt},
        vline{2,Z} = {0.5pt},
        row{1}     = {font=\bfseries},
        cell{1}{2-4} = {bg=lightgray!20},
        cell{2-4}{1} = {bg=lightgray!20},
        column{1}  = {font=\bfseries},
        rows       = {1.5em,m},
        colspec    = {lccc},
    }
     & {Generalization\\ Errors} & {OOD\\ Errors} & {Adverserial\\ Attacks} \\
    Monitoring Inputs  & $\checkmark$ & $\checkmark$ & $\checkmark$ \\
    Monitoring Internal Representations & $\checkmark\checkmark$ & $\checkmark\checkmark\checkmark$ & $\checkmark\checkmark$ \\
    Monitoring Outpus & $\checkmark$ & $\checkmark\checkmark$ & $\checkmark$
    \end{tblr}
  \label{tab:sota_mapping}
\end{table*}

\subsection{Monitoring Inputs}
Input monitoring approaches focus on analyzing the data fed into the neural network to detect potential safety concerns before they propagate through the system. These methods often employ signal
processing techniques or reconstruction models to identify anomalies
in the input space.

Liu et al. introduce a runtime monitor for detecting adversarial examples in DNNs by leveraging local robustness verification. Their approach monitors the robustness radius of network inputs and successfully distinguishes between valid and adversarial samples. Their empirical results on CNNs trained on MNIST~\cite{mnist} and CIFAR-10~\cite{cifar10} demonstrate that adversarial examples have significantly smaller robustness radii compared to correctly classified inputs~\cite{Liu2023InputVF}.
In~\cite{asad2024beyond}, Asad et al. proposes an adversarial autoencoder for novelty detection by estimating the probability distribution of inlier data. Their approach monitors the reconstruction error and latent space distribution, effectively detecting OOD samples in the MNIST, Fashion-MNIST~\cite{fashionmnist}, and Coil-100~\cite{coil100} datasets.

\subsection{Monitoring Internal Representations}

Internal monitoring methods operate by observing the activations within a neural network during inference, leveraging intermediate features and hidden representations to analyze the network's behavior and detect anomalies or novel inputs.
Henzinger et al. use an abstraction-based approach in which they detect OOD inputs during runtime by monitoringe neurons in the hidden layers. They use hyperrectangles for each neuron in the hidden layer to capture the range of activation values observed during training. The monitor then observes the values at runtime and flaggs inputs as OOD if their hidden activation value falls outside the hyperrectangles~\cite{Henzinger2019OutsideTB}.
Lukina et al. propose an monitoring approach to observe feature layer activations~\cite{Lukina2020IntoTU}. These activations are compared against clusters formed during training to detect outliers. It assings confidence scores to predictions, enabling detection of OOD inputs. Moreover, the framework incorporates a human-in-the-loop labeling process to address unknown scenarios and update the neural network and the monitor. Image classification experiments on the MNIST and CIFAR-10~\cite{cifar10} datasets show high runtime accuracy compared to non-adaptive methods. 
Geissler et al. proposed a method that extracts activation distribution quantiles and uses decision trees to classify deviations as potential silent errors in object detection tasks \cite{Geissler2023ALS}. The technique is evaluated on the COCO~\cite{Lin2014Coco} and KITTI \cite{kitti} datasets.

A runtime monitoring framework for 3D object detection in automated driving systems is proposed by Yatbaz et al., focusing on detecting missed objects by analyzing neural activation patterns from multiple layers of the detector's backbone network \cite{Yatbaz2024RuntimeMO}. The method leverages sparse LiDAR data and demonstrates that earlier layers improve error detection compared to final-layer activations. The authors validate their approach on KITTI and NuScenes \cite{nuscenes} datasets using PointPillars \cite{lang2019pointpillars} and CenterPoint \cite{yin2021center} detectors. While not explicitly targeting OOD detection or adversarial attacks, the method addresses performance degradation in complex 3D scenarios, monitoring activation maps to enhance safety.

A recent approach by Wu et al. proposes Box Abstraction Monitors (BAM) to detect OOD samples in object detection. BAM forms non-convex decision boundaries in the feature space using convex box abstractions. Their evaluation on Faster R-CNN-based models shows strong performance on the KITTI and BDD100K \cite{yu2020bdd100k} datasets with minimal computational overhead \cite{Wu2024BAMBA}.

\subsection{Monitoring Outputs}

Output-based runtime safety monitoring focuses on analyzing the predictions
generated by a DNN to detect unsafe behaviors. These methods are particularly practical for deployment, as they require no access to the DNN's internal layers or training data. A prominent approach in this category is uncertainty estimation, which quantifies the confidence of predictions to identify potential errors. For instance, Monte Carlo dropout leverages stochastic forward passes during inference to estimate prediction variance, enabling the detection of misclassifications in tasks like image classification on datasets such as CIFAR-10 \cite{cifar10}. Similarly, softmax entropy analysis serves as a simple yet effective tool.
Kumura et al. applied this principle to adversarial attack detection by measuring the Kullback-Leibler-Divergence between outputs before and after input transformations, demonstrating robustness on benchmarks \cite{kantaros2021real}.

\subsection{Combined Approaches}

Combined approaches in runtime monitoring leverage multiple techniques to address diverse error types by integrating complementary monitoring methods. These approaches often employ meta-models or ensemble strategies to aggregate outputs from individual monitors to cover different failure types.

Klingner et al. propose an adversarial perturbation detection framework for multi-task perception in depth estimation and semantic segmentation. Their approach monitors edge consistency between input images, depth outputs, and segmentation outputs, identifying adversarial perturbations. The detection method is tested on the Cityscapes and KITTI datasets, demonstrating robustness against various attacks \cite{klingner2022detecting}.

\cite{hacker2023insufficiency} propose an insufficiency-driven DNN error detection framework for traffic sign recognition. The method combines multiple runtime monitors targeting distinct DNN insufficiencies such as robustness, interpretability and uncertainty. A meta-model aggregates outputs from individual monitors to predict triggering conditions. The approach is validated using self-generated 3D driving scenarios and public datasets, i.e. GTSDB and GTSRB. The baseline DNN for classification is a 5-layer CNN, while detection uses YOLOv2. Monitored aspects include adversarial perturbations, confidence scores, and saliency maps \cite{hacker2023insufficiency}.

\section{Discussion}\label{sec:discussion}

Runtime safety monitoring plays an important role for deploying deep neural networks in safety-critical perception systems. This survey highlights the breadth of techniques designed to address functional insufficiencies such as adversarial attacks or out-of-distribution inputs. A central observation is that no single runtime safety monitor can comprehensively address all safety risks. Methods that analyze inputs are effective for obvious anomalies but struggle with subtler issues, such as rare environmental conditions. Techniques that inspect internal network activations provide deeper insights into the causes of errors, such as unexpected patterns in hidden layers, but often require high computational resources. Output-based methods, which assess prediction confidence or consistency, offer simplicity but are limited by the DNN's tendency to produce highly confident yet incorrect results in unfamiliar scenarios. These limitations highlight the need for integrated frameworks that combine multiple monitoring strategies.
As DNNs become integral to autonomous systems, runtime safety monitoring will play a pivotal role in ensuring their safe deployment. Future advancements must prioritize adaptability, efficiency, and explainability, ensuring that monitors can detect safety concerns in real-time.

\section*{Acknowledgment}

This work was supported by funding from the Topic Engineering Secure Systems of the Helmholtz Association (HGF) and by KASTEL Security Research Labs (46.23.03).

\bibliographystyle{IEEEtran}
\bibliography{references.bib}

@inproceedings{roy_runtime_2022,
	title = {Runtime Monitoring of Deep Neural Networks Using Top-Down Context Models Inspired by Predictive Processing and Dual Process Theory},
	url = {https://www.semanticscholar.org/paper/Runtime-Monitoring-of-Deep-Neural-Networks-Using-by-Roy-Cobb/7140d2f91a63c8450b4ffde5d39d167e7b26ac28},
	author = {Roy, Anirban and Cobb, Adam D. and Bastian, Nathaniel D. and Jalaian, Brian and Jha, Susmit},
	urldate = {2024-04-08},
	date = {2022},
}

@article{kaur_predicting_2023,
	title = {Predicting Out-of-Distribution Performance of Deep Neural Networks Using Model Conformance},
	rights = {https://doi.org/10.15223/policy-029},
	url = {https://ieeexplore.ieee.org/document/10207589/},
	doi = {10.1109/ICAA58325.2023.00011},
	pages = {19--28},
	journaltitle = {2023 {IEEE} International Conference on Assured Autonomy ({ICAA})},
	author = {Kaur, Ramneet and Jha, Susmit and Roy, Anirban and Sokolsky, Oleg and Lee, Insup},
	urldate = {2024-04-08},
	date = {2023-06},
	note = {Conference Name: 2023 {IEEE} International Conference on Assured Autonomy ({ICAA})
{ISBN}: 9798350326017
Place: Laurel, {MD}, {USA}
Publisher: {IEEE}},
}

@inproceedings{cheng2019runtime,
  title={Runtime monitoring neuron activation patterns},
  author={Cheng, Chih-Hong and N{\"u}hrenberg, Georg and Yasuoka, Hirotoshi},
  booktitle={2019 Design, Automation \& Test in Europe Conference \& Exhibition (DATE)},
  pages={300--303},
  year={2019},
  organization={IEEE}
}

@inproceedings{kantaros2021real,
  title={Real-time detectors for digital and physical adversarial inputs to perception systems},
  author={Kantaros, Yiannis and Carpenter, Taylor and Sridhar, Kaustubh and Yang, Yahan and Lee, Insup and Weimer, James},
  booktitle={Proceedings of the ACM/IEEE 12th International Conference on Cyber-Physical Systems},
  pages={67--76},
  year={2021}
}

@article{Geissler2023ALS,
  title={A Low-Cost Strategic Monitoring Approach for Scalable and Interpretable Error Detection in Deep Neural Networks},
  author={Florian Geissler and Syed Sha Qutub and Michael Paulitsch and Karthik Pattabiraman},
  journal={ArXiv},
  year={2023},
  volume={abs/2310.20349},
  url={https://api.semanticscholar.org/CorpusID:261895249}
}

@inproceedings{Lin2014Coco,
  title={Microsoft COCO: Common Objects in Context},
  author={Tsung-Yi Lin and Michael Maire and Serge J. Belongie and James Hays and Pietro Perona and Deva Ramanan and Piotr Doll{\'a}r and C. Lawrence Zitnick},
  booktitle={European Conference on Computer Vision},
  year={2014},
  url={https://api.semanticscholar.org/CorpusID:14113767}
}

@article{Henzinger2019OutsideTB,
  title={Outside the Box: Abstraction-Based Monitoring of Neural Networks},
  author={Thomas A. Henzinger and Anna Lukina and Christian Schilling},
  journal={ArXiv},
  year={2019},
  volume={abs/1911.09032},
  url={https://api.semanticscholar.org/CorpusID:208175510}
}

@article{Shao2020IncreasingTT,
  title={Increasing the Trustworthiness of Deep Neural Networks via Accuracy Monitoring},
  author={Zhihui Shao and Jianyi Yang and Shaolei Ren},
  journal={ArXiv},
  year={2020},
  volume={abs/2007.01472},
  url={https://api.semanticscholar.org/CorpusID:220347137}
}

@inproceedings{Yatbaz2024RuntimeMO,
  title={Run-time Monitoring of 3D Object Detection in Automated Driving Systems Using Early Layer Neural Activation Patterns},
  author={Hakan Yekta Yatbaz and Mehrdad Dianati and Konstantinos Koufos and Roger Woodman},
  year={2024},
  url={https://api.semanticscholar.org/CorpusID:269042756}
}

@article{Lukina2020IntoTU,
  title={Into the unknown: Active monitoring of neural networks},
  author={Anna Lukina and Christian Schilling and Thomas A. Henzinger},
  journal={ArXiv},
  year={2020},
  volume={abs/2009.06429},
  url={https://api.semanticscholar.org/CorpusID:221655848}
}

@incollection{houben2022inspect,
  title={Inspect, understand, overcome: A survey of practical methods for AI safety},
  author={Houben, Sebastian and Abrecht, Stephanie and Akila, Maram and B{\"a}r, Andreas and Brockherde, Felix and Feifel, Patrick and Fingscheidt, Tim and Gannamaneni, Sujan Sai and Ghobadi, Seyed Eghbal and Hammam, Ahmed and others},
  booktitle={Deep Neural Networks and Data for Automated Driving: Robustness, Uncertainty Quantification, and Insights Towards Safety},
  year={2022},
  publisher={Springer}
}

@article{rahman2021runtime,
  author       = {Quazi Marufur Rahman and
                  Peter Corke and
                  Feras Dayoub},
  title        = {Run-Time Monitoring of Machine Learning for Robotic Perception: {A} Survey of Emerging Trends},
  journal      = {{IEEE} Access},
  volume       = {9},
  year         = {2021},
}

@article{muhammad2020deep,
  title={Deep learning for safe autonomous driving: Current challenges and future directions},
  author={Muhammad, Khan and Ullah, Amin and Lloret, Jaime and Del Ser, Javier and de Albuquerque, Victor Hugo C},
  journal={IEEE Transactions on Intelligent Transportation Systems},
  volume={22},
  number={7},
  pages={4316--4336},
  year={2020},
  publisher={IEEE}
}

@phdthesis{ferreira2023runtime,
  title={Runtime safety monitoring of ML-based perception functions in autonomous systems},
  author={Ferreira, Raul Sena},
  year={2023},
  school={Universit{\'e} Paul Sabatier-Toulouse III}
}

@article{ferreira2024safety,
  author       = {Raul Sena Ferreira and
                  Joris Gu{\'{e}}rin and
                  Kevin Delmas and
                  J{\'{e}}r{\'{e}}mie Guiochet and
                  H{\'{e}}l{\`{e}}ne Waeselynck},
  title        = {Safety Monitoring of Machine Learning Perception Functions: a Survey},
  journal      = {CoRR},
  volume       = {abs/2412.06869},
  year         = {2024},
}

@article{hacker2023insufficiency,
  title={Insufficiency-driven DNN error detection in the context of SOTIF on traffic sign recognition use case},
  author={Hacker, Lukas and Seewig, J{\"o}rg},
  journal={IEEE Open Journal of Intelligent Transportation Systems},
  volume={4},
  pages={58--70},
  year={2023},
  publisher={IEEE}
}

@inproceedings{Pavlitskaya2022AdversarialVO,
  title={Adversarial Vulnerability of Temporal Feature Networks for Object Detection},
  author={Svetlana Pavlitskaya and Nikolai Polley and Michael Weber and Johann Marius Z{\"o}llner},
  booktitle={ECCV Workshops},
  year={2022},
  url={https://api.semanticscholar.org/CorpusID:251741092}
}

@inproceedings{pavlitskaya2022suppress,
  author    = {Svetlana Pavlitskaya and
               Jonas Hendl and
               Sebastian Kleim and
               Leopold M{\"{u}}ller and
               Fabian Wylczoch and
               J. Marius Z{\"{o}}llner},
  title     = {Suppress with a Patch: Revisiting Universal Adversarial Patch Attacks against Object Detection},
  booktitle = {{IEEE} International Conference on Electrical, Computer, Communications and Mechatronics Engineering (ICECCME)},
  publisher = {{IEEE}},
  year      = {2022},
}

@article{Pavlitskaya2020FeasibilityAS,
  title={Feasibility and Suppression of Adversarial Patch Attacks on End-to-End Vehicle Control},
  author={Svetlana Pavlitskaya and Sefa {\"U}nver and Johann Marius Z{\"o}llner},
  journal={2020 IEEE 23rd International Conference on Intelligent Transportation Systems (ITSC)},
  year={2020},
  pages={1-8},
  url={https://api.semanticscholar.org/CorpusID:229704142}
}

@article{Goodfellow2014ExplainingAH,
  title={Explaining and Harnessing Adversarial Examples},
  author={Ian J. Goodfellow and Jonathon Shlens and Christian Szegedy},
  journal={CoRR},
  year={2014},
  volume={abs/1412.6572},
  url={https://api.semanticscholar.org/CorpusID:6706414}
}

@article{Szegedy2013IntriguingPO,
  title={Intriguing properties of neural networks},
  author={Christian Szegedy and Wojciech Zaremba and Ilya Sutskever and Joan Bruna and D. Erhan and Ian J. Goodfellow and Rob Fergus},
  journal={CoRR},
  year={2013},
  volume={abs/1312.6199},
  url={https://api.semanticscholar.org/CorpusID:604334}
}

@inproceedings{pavlitska2023adversarial,
	author       = {Svetlana Pavlitska and
                  Nico Lambing and
                  J. Marius Z{\"{o}}llner},
    title        = {Adversarial Attacks on Traffic Sign Recognition: {A} Survey},
    booktitle = "International Conference on Electrical, Computer, Communications and Mechatronics Engineering (ICECCME)",
    year = 2023
}

@article{Madry2017TowardsDL,
  title={Towards Deep Learning Models Resistant to Adversarial Attacks},
  author={Aleksander Madry and Aleksandar Makelov and Ludwig Schmidt and Dimitris Tsipras and Adrian Vladu},
  journal={ArXiv},
  year={2017},
  volume={abs/1706.06083},
  url={https://api.semanticscholar.org/CorpusID:3488815}
}

@inproceedings{willers2020safety,
  title={Safety concerns and mitigation approaches regarding the use of deep learning in safety-critical perception tasks},
  author={Willers, Oliver and Sudholt, Sebastian and Raafatnia, Shervin and Abrecht, Stephanie},
  booktitle={Computer Safety, Reliability, and Security. SAFECOMP 2020 Workshops: DECSoS 2020, DepDevOps 2020, USDAI 2020, and WAISE 2020, Lisbon, Portugal, September 15, 2020, Proceedings 39},
  pages={336--350},
  year={2020},
  organization={Springer}
}

@techreport{sotif,
  address = {Geneva, CH},
  type = {Standard},
  title = {Road vehicles -- {Safety} of the intended functionality},
  shorttitle = {{ISO} 21448:2022},
  url = {https://www.iso.org/standard/77490.html},
  language = {en},
  number = {ISO 21448:2022},
  institution = {International Organization for Standardization},
  author = {{ISO Central Secretary}},
  year = {2022}
}

@article{Jin2019QuantifyingTG,
  title={Quantifying the generalization error in deep learning in terms of data distribution and neural network smoothness},
  author={Pengzhan Jin and Lu Lu and Yifa Tang and George Em Karniadakis},
  journal={Neural networks : the official journal of the International Neural Network Society},
  year={2019},
  volume={130},
  pages={
          85-99
        },
  url={https://api.semanticscholar.org/CorpusID:263796258}
}

@article{Jakubovitz2018GeneralizationEI,
  title={Generalization Error in Deep Learning},
  author={Daniel Jakubovitz and Raja Giryes and Miguel R. D. Rodrigues},
  journal={ArXiv},
  year={2018},
  volume={abs/1808.01174},
  url={https://api.semanticscholar.org/CorpusID:51921368}
}

@article{hendrycks2016baseline,
  title={A baseline for detecting misclassified and out-of-distribution examples in neural networks},
  author={Hendrycks, Dan and Gimpel, Kevin},
  journal={arXiv preprint arXiv:1610.02136},
  year={2016}
}

@article{liang2017enhancing,
  title={Enhancing the reliability of out-of-distribution image detection in neural networks},
  author={Liang, Shiyu and Li, Yixuan and Srikant, Rayadurgam},
  journal={arXiv preprint arXiv:1706.02690},
  year={2017}
}

@inproceedings{Hashemi2022RuntimeMF,
  title={Runtime Monitoring for Out-of-Distribution Detection in Object Detection Neural Networks},
  author={Vahid Moraveji Hashemi and Jan Křet{\'i}nsk{\'y} and Sabine Rieder and Jessica Schmidt},
  booktitle={World Congress on Formal Methods},
  year={2022},
  url={https://api.semanticscholar.org/CorpusID:254685917}
}

@article{Hell2021MonitoringPR,
  title={Monitoring perception reliability in autonomous driving: Distributional shift detection for estimating the impact of input data on prediction accuracy},
  author={Franz Hell and Gereon Hinz and Feng Liu and Sakshi Goyal and Ke Pei and Tetiana Lytvynenko and Alois Knoll and Yiqiang Chen},
  journal={Proceedings of the 5th ACM Computer Science in Cars Symposium},
  year={2021},
  url={https://api.semanticscholar.org/CorpusID:244663680}
}

@article{Wu2024BAMBA,
  title={BAM: Box Abstraction Monitors for Real-time OoD Detection in Object Detection},
  author={Changshun Wu and Weicheng He and Chih-Hong Cheng and Xiaowei Huang and Saddek Bensalem},
  journal={ArXiv},
  year={2024},
  volume={abs/2403.18373},
  url={https://api.semanticscholar.org/CorpusID:268723632}
}

@article{kitti,
  title={Vision meets robotics: The kitti dataset},
  author={Geiger, Andreas and Lenz, Philip and Stiller, Christoph and Urtasun, Raquel},
  journal={The International Journal of Robotics Research},
  volume={32},
  number={11},
  pages={1231--1237},
  year={2013},
  publisher={Sage Publications Sage UK: London, England}
}

@inproceedings{nuscenes,
  title={nuscenes: A multimodal dataset for autonomous driving},
  author={Caesar, Holger and Bankiti, Varun and Lang, Alex H and Vora, Sourabh and Liong, Venice Erin and Xu, Qiang and Krishnan, Anush and Pan, Yu and Baldan, Giancarlo and Beijbom, Oscar},
  booktitle={Proceedings of the IEEE/CVF conference on computer vision and pattern recognition},
  pages={11621--11631},
  year={2020}
}

@inproceedings{lang2019pointpillars,
  title={Pointpillars: Fast encoders for object detection from point clouds},
  author={Lang, Alex H and Vora, Sourabh and Caesar, Holger and Zhou, Lubing and Yang, Jiong and Beijbom, Oscar},
  booktitle={Proceedings of the IEEE/CVF conference on computer vision and pattern recognition},
  pages={12697--12705},
  year={2019}
}

@inproceedings{yin2021center,
  title={Center-based 3d object detection and tracking},
  author={Yin, Tianwei and Zhou, Xingyi and Krahenbuhl, Philipp},
  booktitle={Proceedings of the IEEE/CVF conference on computer vision and pattern recognition},
  pages={11784--11793},
  year={2021}
}

@inproceedings{yu2020bdd100k,
  title={Bdd100k: A diverse driving dataset for heterogeneous multitask learning},
  author={Yu, Fisher and Chen, Haofeng and Wang, Xin and Xian, Wenqi and Chen, Yingying and Liu, Fangchen and Madhavan, Vashisht and Darrell, Trevor},
  booktitle={Proceedings of the IEEE/CVF conference on computer vision and pattern recognition},
  pages={2636--2645},
  year={2020}
}

@inproceedings{klingner2022detecting,
  title={Detecting adversarial perturbations in multi-task perception},
  author={Klingner, Marvin and Kumar, Varun Ravi and Yogamani, Senthil and B{\"a}r, Andreas and Fingscheidt, Tim},
  booktitle={2022 IEEE/RSJ International Conference on Intelligent Robots and Systems (IROS)},
  pages={13050--13057},
  year={2022},
  organization={IEEE}
}

@article{cifar10,
  title={Learning multiple layers of features from tiny images},
  author={Krizhevsky, Alex and Hinton, Geoffrey and others},
  year={2009},
  publisher={Toronto, ON, Canada}
}

@article{mnist,
  title={MNIST handwritten digit database},
  author={LeCun, Yann and Cortes, Corinna and Burges, CJ},
  journal={ATT Labs [Online]. Available: http://yann.lecun.com/exdb/mnist},
  volume={2},
  year={2010}
}

@article{Liu2023InputVF,
  title={Input Validation for Neural Networks via Local Robustness Verification},
  author={Jiangchao Liu and Liqian Chen and Antoine Min{\'e} and Hengbiao Yu and Ji Wang},
  journal={2023 IEEE 23rd International Conference on Software Quality, Reliability, and Security Companion (QRS-C)},
  year={2023},
  pages={237-246},
  url={https://api.semanticscholar.org/CorpusID:267771461}
}

@article{asad2024beyond,
  title={Beyond the Known: Adversarial Autoencoders in Novelty Detection},
  author={Asad, Muhammad and Ullah, Ihsan and Sistu, Ganesh and Madden, Michael G},
  journal={arXiv preprint arXiv:2404.04456},
  year={2024}
}

@inproceedings{coil100,
  title={Columbia Object Image Library (COIL100)},
  author={Sameer A. Nene and Shree K. Nayar and Hiroshi Murase},
  year={1996},
  url={https://api.semanticscholar.org/CorpusID:58758670}
}

@online{fashionmnist,
  author       = {Han Xiao and Kashif Rasul and Roland Vollgraf},
  title        = {Fashion-MNIST: a Novel Image Dataset for Benchmarking Machine Learning Algorithms},
  date         = {2017-08-28},
  year         = {2017},
  eprintclass  = {cs.LG},
  eprinttype   = {arXiv},
  eprint       = {cs.LG/1708.07747},
}

@article{huang2020asruvey,
  author       = {Xiaowei Huang and
                  Daniel Kroening and
                  Wenjie Ruan and
                  James Sharp and
                  Youcheng Sun and
                  Emese Thamo and
                  Min Wu and
                  Xinping Yi},
  title        = {A survey of safety and trustworthiness of deep neural networks: Verification,
                  testing, adversarial attack and defence, and interpretability},
  journal      = {Comput. Sci. Rev.},
  year         = {2020},
}

@article{mohseni2023taxonomy,
  author       = {Sina Mohseni and
                  Haotao Wang and
                  Chaowei Xiao and
                  Zhiding Yu and
                  Zhangyang Wang and
                  Jay Yadawa},
  title        = {Taxonomy of Machine Learning Safety: {A} Survey and Primer},
  journal      = {{ACM} Comput. Surv.},
  year         = {2023},
}

@article{yatbaz2024introspection,
  author       = {Hakan Yekta Yatbaz and
                  Mehrdad Dianati and
                  Roger Woodman},
  title        = {Introspection of DNN-Based Perception Functions in Automated Driving
                  Systems: State-of-the-Art and Open Research Challenges},
  journal      = {{IEEE} Trans. Intell. Transp. Syst.},
  year         = {2024},
}

@article{hussain2022deepguard,
  title={DeepGuard: A framework for safeguarding autonomous driving systems from inconsistent behaviour},
  author={Hussain, Manzoor and Ali, Nazakat and Hong, Jang-Eui},
  journal={Automated Software Engineering},
  volume={29},
  number={1},
  pages={1},
  year={2022},
  publisher={Springer}
}

@inproceedings{wang2020dnnguard,
  title={Dnnguard: An elastic heterogeneous dnn accelerator architecture against adversarial attacks},
  author={Wang, Xingbin and Hou, Rui and Zhao, Boyan and Yuan, Fengkai and Zhang, Jun and Meng, Dan and Qian, Xuehai},
  booktitle={Proceedings of the Twenty-Fifth International Conference on Architectural Support for Programming Languages and Operating Systems},
  pages={19--34},
  year={2020}
}

\end{document}